\documentclass{svproc}
\usepackage{url}
\usepackage{graphicx}
\usepackage{multicol}
\usepackage{caption}
\usepackage{subcaption}
\usepackage{todonotes}
\usepackage{color, colortbl}
\usepackage{rotating}
\usepackage{multirow}
\usepackage{longtable}
\usepackage{adjustbox}
\usepackage{mathtools}
\usepackage{amssymb}
\usepackage{amsmath}
\usepackage{enumerate}
\usepackage{enumitem}
\usepackage{fancyvrb}

\usepackage{siunitx} 

\usepackage[backend=biber,style=nature, doi=true, url=true]{biblatex}

\DeclareBibliographyDriver{article}{%
  \usebibmacro{bibindex}%
  \usebibmacro{begentry}%
  \usebibmacro{author/translator+others}%
  \setunit{\labelnamepunct}\newblock
  \iftoggle{bbx:articletitle}
    {%
      \usebibmacro{title}%
      \newblock
    }
    {}%
  \printlist{language}%
  \newunit\newblock
  \usebibmacro{byauthor}%
  \newunit\newblock
  \usebibmacro{bytranslator+others}%
  \newunit\newblock
  \printfield{version}%
  \newunit\newblock
  \usebibmacro{journal+issuetitle}%
  \newunit
  \usebibmacro{byeditor+others}%
  \newunit
  \usebibmacro{note+pages}%
  \newunit
  \iftoggle{bbx:isbn}
    {\printfield{issn}}
    {}%
  \newunit\newblock
  \usebibmacro{doi+eprint+url}%
  \setunit{\addspace}\newblock
  \usebibmacro{issue+date}%
  \newunit\newblock
  \usebibmacro{addendum+pubstate}%
  \setunit{\bibpagerefpunct}\newblock
  \usebibmacro{pageref}%
  \newunit\newblock
  \usebibmacro{related}%
  \usebibmacro{finentry}%
}

\begin{document}
\mainmatter              
\title{Power to the springs: Passive elements are sufficient to drive push-off in human walking}
\titlerunning{Power to the springs} 

\author{Alexandra Buchmann\inst{1} \and Bernadett Kiss\inst{2} \and Alexander Badri-Spr{\"{o}}witz\inst{2} \and Daniel Renjewski\inst{1}}
\authorrunning{Alexandra Buchmann et al.} 
\tocauthor{Alexandra Buchmann, Bernadett Kiss, Alexander Badri-Spr{\"{o}}witz, Daniel Renjewski}
\institute{Chair of Applied Mechanics, Department of Mechanical Engineering, TUM School of Engineering and Design, Technical University Munich, Boltzmannstr. 15, 85748 Garching, Germany\\
\email{alexandra.buchmann@tum.de}\\ 
WWW homepage:
\texttt{https://www.mec.ed.tum.de/am/}
\and
Dynamic Locomotion Group, Max Planck Institute for Intelligent Systems,  Heisenbergstr. 3, 70569 Stuttgart, Germany}

\maketitle              

\begin{abstract}
For the impulsive ankle push-off (APO) observed in human walking two muscle-tendon-units (MTUs) spanning the ankle joint play an important role: Gastrocnemius (GAS) and Soleus (SOL). GAS and SOL load the Achilles tendon to store elastic energy during stance followed by a rapid energy release during APO. 
We use a neuromuscular simulation (NMS) and a bipedal robot to investigate the role of GAS and SOL on the APO. We optimize the simulation for a robust gait and then sequentially replace the MTUs of (1)\,GAS, (2)\,SOL and (3)\,GAS\,and\,SOL by linear springs. To validate the simulation, we implement NMS-3 on a bipedal robot.
Simulation and robot walk steady for all trials showing an impulsive APO. Our results imply that the elastic MTU properties shape the impulsive APO. For prosthesis or robot design that is, no complex ankle actuation is needed to obtain an impulsive APO, if more mechanical intelligence is incorporated in the design. 

\end{abstract}

\section{Introduction}%
Plantigrade bipedal walking, i.e. walking with heel and toes flat on the ground, distinguishes humans from most other animals. The complexity of human bipedal gait becomes visible when trying to technically replicate or restore natural human leg dynamics. Humanoid robots, lower limb prostheses or exoskeletons fall far short of human performance in terms of efficiency, versatility and robustness \cite{Sygulla.2020, Lechler.2018, Cardona.2020}. This can be partially attributed to technical limitations, but more importantly to a lack of understanding the underlying biomechanical and control principles of human walking.\par
%
A characteristic feature of human walking is the high power output at the ankle joint in late stance \cite{Perry.2010} which leads to an impulsive ankle push-off (APO) \cite{Meinders.1998}. It has been shown that the peak power is substantially higher than maximum active power capacity of the plantar flexor muscle's Gastrocnemius (GAS) and Soleus (SOL) \cite{Lipfert.2014}. The power peak higher than the muscles' capacity indicates the use of elastic structures to store and rapidly release mechanical energy contributing to the energy efficiency of locomotion \cite{Alexander.1977, Alexander.1991, Alexander.1992}.\par
%
\textcite{Hof.1983} were the first to label the impulsive APO in human walking a catapult. 
Catapult mechanisms are commonly used during locomotion in nature. Humans use a catapult during APO to propel their trailing leg into swing \cite{Lipfert.2014}, as do galloping horses which use a catapult to achieve rapid limb protraction \cite{Wilson.2003}. Locusts, frogs, fleas, and click beetles use a geometrical catapult for jumping \cite{Gronenberg.1996, Clark.1975, Nishikawa.1999}.\\
Functionally, a catapult is characterized by slow storage of elastic energy followed by a rapid energy release with substantially higher power to accelerate a projectile. Three main components are needed to achieve this power amplification: an elastic element to store energy, a block or frame to take up forces arising during loading and a catch to hold the elastic energy and trigger its rapid release.\par
%
In all biological examples, elastic energy storage is facilitated by muscles-tendon-units (MTUs). During human APO, SOL and GAS MTUs provide up to 91\% of the power from elastic energy \cite{Cronin.2013}. During stance, GAS and SOL mostly work isometrically \cite{Ishikawa.2005} while the Achilles tendon is stretched to store energy. The MTUs of GAS and SOL thus act like springs reducing the metabolic costs required for walking \cite{Cronin.2013}.\\
%
The observation that ankle kinematics and kinetics in human walking are mainly driven by the passive-elastic properties of ankle plantar flexor MTUs raises the following questions: 
\begin{enumerate}[topsep=2pt,itemsep=5pt,parsep=0pt,partopsep=0pt]
\item[I] Do humans need active ankle actuation for steady state walking with impulsive ankle push-off or would passive-elastic structures be sufficient?
\item[II] What are the individual contributions of the plantar flexor muscle-tendon-units, Soleus and Gastrocnemius, to the ankle push-off?
\end{enumerate}
We investigate (I) and (II) using a 2D neuromuscular simulation (NMS) of human walking \cite{Geyer.2010} and a bipedal robot for real world validation. First, we optimize the simulation for a robust, natural gait and then replace (1)\,GAS, (2)\,SOL and (3)\,GAS\,and\,SOL MTUs by passive linear springs without adapting neural control parameters. Finally, we implement NMS-3 on a bipedal robot with actuated knee and hip joints and passive elastic ankle joint, where springs connected to cables spanning the ankle joint represent GAS and SOL MTUs. NMS and robot were able to walk continuously in all trials showing an impulsive APO even in absence of active ankle actuation.\par
For prosthesis, exoskeleton, or bipedal robot design we state that more attention should be payed to an intelligent mechanical ankle design inspired by human morphology and biomechanical function. Reproducing natural leg dynamics with impulsive APO may be possible by utilizing passive-elastic elements for steady state bipedal walking instead of complex control or powerful actuator strategies.

\section{Methods}
\subsection{Neuromuscular Simulation and Solver}
\begin{figure}[h]
    \begin{subfigure}[c]{0.245\textwidth}
    \centering
    \includegraphics[trim=0cm 0cm 15cm 0.5cm, clip, height=4cm]{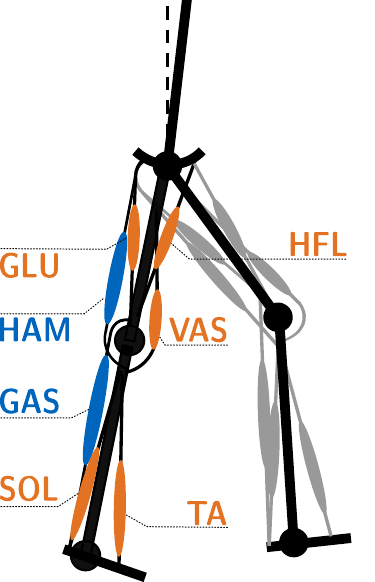}
    \subcaption{NMS-ref.}
    \end{subfigure}
    \begin{subfigure}[c]{0.245\textwidth}
    \includegraphics[trim=4.45cm 0cm 10cm 0.5cm, clip, height=4cm]{model.pdf}
    \subcaption{NMS-1.}
    \end{subfigure}
    \begin{subfigure}[c]{0.245\textwidth}
    \includegraphics[trim=9.45cm 0cm 5cm 0.5cm, clip, height=4cm]{model.pdf}
    \subcaption{NMS-2.}
    \end{subfigure}
    \begin{subfigure}[c]{0.245\textwidth}
    \includegraphics[trim=14.5cm 0cm 0cm 0.5cm, clip, height=4cm]{model.pdf}
    \subcaption{NMS-3.}
    \end{subfigure}
    \caption[Simulation experiments and model setup.]{\textbf{Simulation experiments and model setup.} (a) NMS-ref configuration with seven muscle groups per leg namely Gluteus \textit{GLU}, Hamstrings \textit{HAM}, Gastrocnemius \textit{GAS}, Soleus \textit{SOL}, Tibialis Anterior \textit{TA}, Vastus \textit{VAS} and Hip flexor muscles \textit{HFL}. For details on the NMS see \cite{Geyer.2010}. (b) to (d): three NMS trials where (1)\,GAS, (2)\,SOL and (3)\,GAS\,and\,SOL MTUs are replaced by linear springs with stiffness $k$ and resting length $l_0$. Figure inspired by \cite{Geyer.2019}.}
    \label{fig:cases}
\end{figure}

We used a 2D neuromuscular simulation \cite{Geyer.2010} with 9 degrees of freedom (trunk-to-world: 3, knee, ankle and hip 1 each) running on Matlab Simulink R2020b (The MathWorks Inc., Natick, Massachusetts, America). The NMS includes seven muscles groups per leg controlled by reflex feedback. Fig.\,\ref{fig:cases}\,(a) shows the setup of the NMS-ref model and Fig.\,\ref{fig:cases}\,(b)-(d) NMS-1 to NMS-3 where we replaced GAS and/or SOL MTUs by springs.\\
The NMS is solved using a variable-order numerical solver \textit{ode15s}\footnote{\url{https://www.mathworks.com/help/simulink/gui/solver.html}} 
for stiff differential equations with a maximum step size of \SI{0.1}{s}, a minimum step size of $1\,e^{-9}$ and relative and absolute tolerances of $1\,e^{-3}$ and $1\,e^{-4}$, respectively. The zero-crossing detection was set to be non-adaptive with a maximum number of 30 consecutive crossings before aborting the simulation.
%
\vspace{-0.1cm}

\vspace{-0.5cm}
\subsection{Optimization}
\label{subsec:optim}
To optimize the simulation for a natural, steady state walking we implemented a global optimization framework using a non-deterministic genetic algorithm \textit{Matlab GA\footnote{\url{https://www.mathworks.com/help/gads/ga.html}}} with a population size of 200 and a maximum of 50 stall generations as termination criterion.
We ran the optimization on a 64-bit computer using a 12-core \textit{Intel i9-10900K} CPU at 3.7\,GHz with 64\,GB RAM in parallel computing and a pre-compiled model in accelerator mode with fast restart.\par
We designed a custom cost function to optimize ten of the model's muscle reflex gains. The resulting gains are given in Tab.\,\ref{tab:params}. Upper and lower bounds were selected according to boundary values for stable gait from \cite[Table I, p.269]{Geyer.2010}.\\
The cost function is inspired by \textcite{Veerkamp.2021} and \textcite{Falisse.2019}. We included the metabolic energy consumption, muscle fatigue expressed by muscle activation squared \cite{Crowninshield.1981}, ground reaction force jerks, the acceleration of the head arm trunk segment (HAT) and knee stop moments representing the passive structures in the knee, e.g. ligaments (Eq.\,\eqref{eq:Cost_Fun}). We additionally included ankle stop moments to ensure the model stays within the physiologically range of motion and does not use ankle end stops. We calculated stop moments, muscle activations and ground reaction force jerks for both legs summed. The metabolic cost calculation follows \cite{Umberger.2003}.

\def\TermGRFs{\left( \left|\frac{d GRF_x}{dt}    \right| + \left|\frac{d GRF_y}{dt}     \right| \right)}
\def\TermVel{\left( \left|\frac{d v_{HAT, x}}{dt}\right| + \left|\frac{d v_{HAT, y}}{dt}\right| \right) }

\begin{multline} 
J = \frac{1}{x_{end}} 
    \int_{t_{step_3}}^{t_{end}}
    \Bigl(\,
    \underbrace{w_1 \cdot E_{Met}    }_{\substack{\text{metabolic}      \\ \text{energy}}}       + 
    \underbrace{w_2 \cdot ACT^2      }_{\substack{\text{muscle}         \\ \text{fatigue}}}      + 
    \underbrace{w_3 \cdot GRF_{jerk} }_{\substack{\text{ground reaction}\\ \text{force jerks}}}  + 
    \underbrace{w_4 \cdot a_{HAT}    }_{\substack{\text{HAT}            \\ \text{acceleration}}} + \, ... \\[2pt]
    ...\,  
    \underbrace{w_5 \cdot |T_{K}|    }_{\substack{\text{stop moments}   \\ \text{knee}}}         +
    \underbrace{w_6 \cdot | T_{A} |  }_{\substack{\text{stop moments}   \\ \text{ankle}}} \,\Bigr) dt \, + 
    \underbrace{w_7 \cdot (t_{max, sim} - t_{end})^2    }_{\substack{\text{early fall}  \\ \text{penalty}}}
    %
\label{eq:Cost_Fun}
\end{multline}
\vspace{0.1cm}

For integration we used the right Riemann summation. All quantities are scaled to the maximum walking distance reached after \SI{10}{s} simulation time. Evaluating steady state walking only, the integral in the cost function was evaluated starting from the third step onwards. If the NMS stopped before finishing three steps the entire simulation time was evaluated to have a smooth reward function also in absence of a good initial guess where the model falls before reaching steady state.
If the NMS aborted before reaching \SI{10}{s} a penalty was applied. The NMS stopped with an error, if the center of mass height fell below \SI{1}{m}, the trunk's forward velocity decreased below zero or joint limits were exceeded reflecting the physiological range of motion.\par
To tune the weights of the cost function we ran the NMS once with the initial parameter set from \textcite{Geyer.2010} and calculated each contribution separately. We then tuned the weights based on the recommendations for relative weighting given in \textcite[Table 3, p.7]{Veerkamp.2021}. For the ankle stop moment we selected a high weight compared to all other values to steer the optimization towards solutions avoiding ankle hard stops. Details see Tab.\,\ref{tab:weights}.

\begin{table}[!htb]
    \begin{minipage}{.475\linewidth}
    \centering
    \caption[Cost function weights and term contributions for NMS-ref]{\textbf{Cost function weights}. $Value$ shows the final weighting, \textit{Rel.\,w} the relative weights used for tuning from \cite{Veerkamp.2021}. $J_{Cont}$ and $\%$ are the absolute contributions after integration for one step and the contribution fractions to the reward function for NMS-ref. \vspace{0.25cm}}
    \setlength{\tabcolsep}{0pt}
    \begin{tabular*}{\columnwidth}{@{\extracolsep{\fill}\quad} l l l l l l}
     Property           & No.    & Value & Rel. w   & $J_{Cont}$    & \%       \\
     \hline
     $E_{CoT}$          & $w_1$ & $5\,e^2$  & 5     & 239           & 69.9    \\
     $ACT^2$            & $w_2$ & $10$      & 0.1   & 2.9           & 0.9     \\
     $GRF_{Jerk}$       & $w_3$ & $2$       & 1     & 40            & 11.7    \\
     $HAT_{acc}$        & $w_4$ & $2\,e^4$  & 1.5   & 45            & 13.2    \\
     $T_{K}$            & $w_5$ & $10$      & 0.25  & 15            & 4.4     \\
     $T_{A}$            & $w_6$ & $1\,e^4$  & n.a.  & 0             & -         \\
     Penalty            & $w_7$ & $1\,e^5$  & n.a.  & 0             & -         \\
     \hline
    \end{tabular*}
    \label{tab:weights}
    \end{minipage}%
    \hfill
    \begin{minipage}{.475\linewidth}
    \centering
    \caption[Optimized Model Reflex Gains.]{\textbf{Muscle reflex gains for NMS-ref.} Muscle gains of force feedback \textbf{(bold)} are normalized to $F_{max}$ of the respective muscle. \vspace{0.25cm}}
    \renewcommand{\arraystretch}{1.1}
    \setlength{\tabcolsep}{0pt}
    \begin{tabular*}{\columnwidth}{@{\extracolsep{\fill}\quad} l l }
        Gain                & NMS-ref\\[2pt]
        \hline 
        $k_{\varphi}$       & 4.25   \\  
        $G_{HAMHFL}$        & 3      \\  
        $\mathbf{G_{GAS}}$  & 1.10    \\ 
        $G_{TA}$            & 1.10 \\
        $\mathbf{G_{VAS}}$  & 1.15 \\ 
        $\mathbf{G_{SOLTA}}$& 0.3  \\
        $\mathbf{G_{SOL}}$  & 1.2  \\
        $G_{HFL}$           & 0.35 \\ 
        $\mathbf{G_{HAM}}$  & 0.65 \\ 
        $\mathbf{G_{GLU}}$  & 0.4  \\        
        \hline
    \end{tabular*}
    \label{tab:params}
    \end{minipage} 
\end{table}

\subsection{Muscle-Tendon-Unit Replacement}
\label{subsec:mtu_repl}
\begin{figure}[htb]
    \centering
    \includegraphics[trim=0cm 0.15cm 0cm 0.4cm, clip, width=0.95\textwidth]{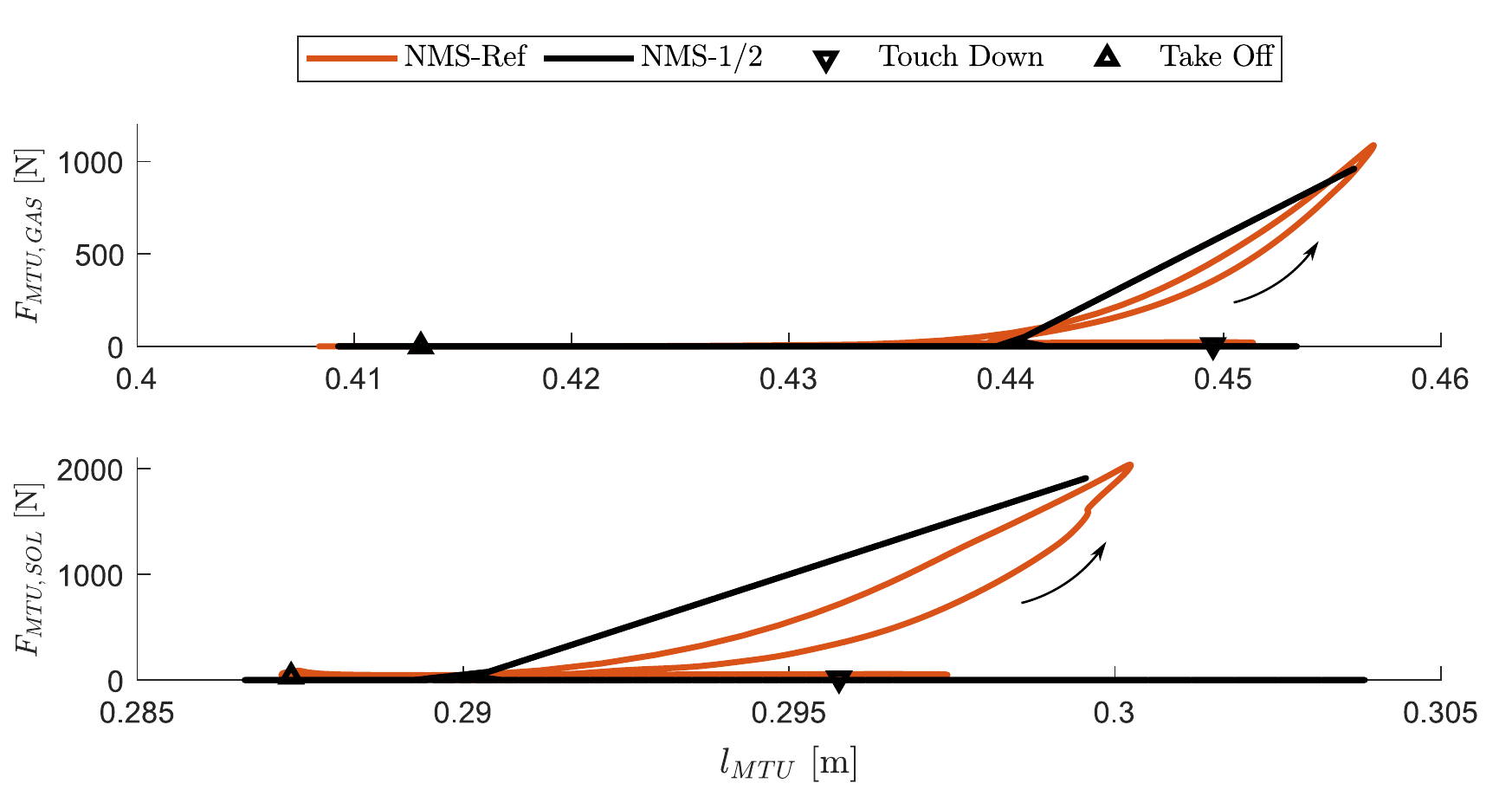}
    \caption[Force length curves of GAS\,\&\,SOL MTUs.]{\textbf{Force length curves of GAS and SOL MTUs during one stride}. \textit{Note the different x- and y-axis for both plots.} The arrow indicates the direction of the orange work loops for the MTUs in NMS-ref for GAS at the top and SOL at the bottom.
    The black lines shows the force length curves when the MTU is replaced by a spring from NMS-1 in the top graph and NMS-2 for the bottom. The triangles denote take off and touch down.
    }
    \label{fig:FL_GAS_SOL}
\end{figure}

For NMS-1,2,3 we replaced the MTUs of GAS and/or SOL by linear mechanical springs as shown in Fig.\,\ref{fig:cases} \textit{(b)-(d)}. To determine spring stiffness and resting length, we analyzed the force-length curves of both MTUs in NMS-ref (Fig.\,\ref{fig:FL_GAS_SOL}).\par
We measured the MTUs length at minimum and maximum force in ascending force-length slope of stance phase and fitted these points coarse with a linear stiffness rounding to the next full ten \SI{}{N/mm} resulting in $l_{0,GAS} = 0.440\,m$, $k_{GAS} = 60\,N/mm$, $l_{0,SOL} = 0.290\,m$, $k_{SOL} = 200\,N/mm$.\\ 
The joint torque resulting from $F_s$ is modeled by $\tau_s = r_s(\varphi)F_S$ with $r_s = r_0 \cos{\varphi - \varphi_{max}}$, $r_0 = 50\,mm$ and $\varphi_{max} = 110^o$ for GAS and SOL at the ankle and $r_0 = 50\,mm$ and $\varphi_{max} = 140^o$ for GAS at the knee \cite[Appendix III]{Geyer.2010}.\par
The spring replacements start loading when heel and ball are on the ground and are instantaneously switch off when the foot leaves the ground. We used the muscle gains given in Tab.\,\ref{tab:params} for all NMS trails, started the model in NMS-ref configuration with MTUs and turned on the springs after the third step when the model was already in a steady state.

\subsection{Robotic Model}
For validation we implemented NMS-3 on a child-size bipedal robot with an anthropomorphic ankle configuration \cite{Kiss.2022}. 
SOL and GAS MTUs are represented by linear springs connected to cables acting on pulleys of fixed radii around the ankle- and knee joints. The SOL and GAS spring stiffness values are \SI{4.5}{N/mm} and \SI{1.4}{N/mm}, respectively with pulley-radii of \SI{13}{mm} at the ankle for SOL and GAS springs, and \SI{13}{mm} for the GAS spring at the knee.\par
The robot's hip and knee joints are actuated and controlled by an open-loop central pattern generator \cite{ijspeert_central_2008} with \SI{1.0}{Hz} locomotion frequency which allows for a walking speed of \SI{0.55}{m/s} on a treadmill. Control patterns roughly follow the NMS joint trajectories. The robot's ankle joints are not actively actuated, only SOL and GAS springs act on them. The designed anthropomorphic bipedal robot weights \SI{2.22}{kg}, at a leg length of \SI{0.35}{m}.\par
The robot is equipped to measure joint angles and velocities. The ankle joint power is calculated from the stiffness values of the SOL and GAS springs when they are loaded according to:
\begin{equation*}
    P_{A}(\alpha, \dot{\varphi}) = 
\begin{cases}
     \dot{\varphi}_A \cdot (k_{SOL} \cdot r^2_{SOL} \cdot \alpha_A + k_{GAS} \cdot r^2_{GAS} \cdot (\alpha_A - \alpha_K)),& \text{if } \alpha_A \geq \alpha_K\\
     \dot{\varphi}_A \cdot k_{SOL} \cdot r^2_{SOL} \cdot \alpha_A,              & \text{otherwise}
\end{cases}
\label{eq:ankle_power_robot}
\end{equation*}
with $P_{A}$ is the ankle power and $\dot{\varphi}_A$ the ankle angular velocity in \SI{}{rad/s}. $\alpha_A$ and $\alpha_K$ are the ankle and knee angle in rad, $k_{SOL}$ and $k_{GAS}$ are the SOL and GAS spring stiffness values in \SI{}{N/m}, and $r_{SOL}$ and $r_{GAS}$ are the pulley radii of the SOL and GAS spring-cables in \SI{}{m}.\\ 
Both springs are loaded starting in plantarflexion at a fixed angle ($-22^o$) as the ankle dorsiflexes. SOL's cable length only depends on the ankle angle while GAS's also depends on the knee angle. \textit{For further details on design and experimental setup see \cite{Kiss.2022}}.

\section{Results}
\subsection{Simulation}
The NMS walks continuously for all experiments. Average forward velocity, stride time and duty factor show minor changes. Metabolic costs decrease from NMS-1 over NMS-2 to NMS-3 compared to NMS-ref by max. $20\%$ for NMS-3 as fewer muscles consume energy when being replaced by a spring (Tab.\,\ref{tab:ankle}).\par

\begin{figure}[htb]
    \centering
    \includegraphics[trim=0.8cm 0.5cm 1.2cm 0.5cm, clip, width=\textwidth]{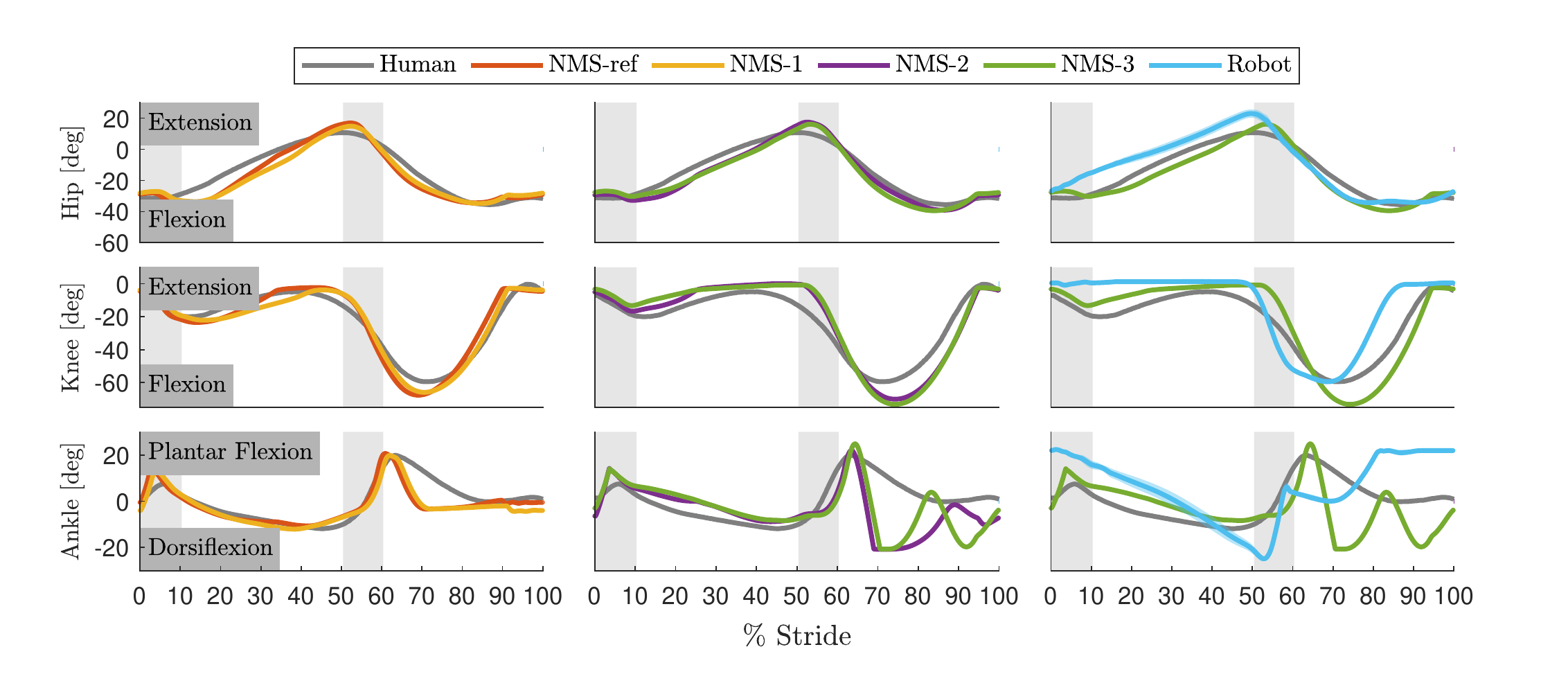}
    \caption[RoM Joints.]{\textbf{Kinematics for hip, knee and ankle.} Human data from \cite{Perry.2010} and sixth step for all NMS trials. Joint extension is always positive. Gray areas indicate human double support. \textit{Left column:} NMS-ref and NMS-1 with very similar trajectories for all joints. \textit{Middle column:} NMS-2 and NMS-3 with replacement of SOL MTU with deviations to human data at ankle and knee joint. \textit{Right column:} NMS-3 and the robot. Knee and hip trajectories of the robot show similar shapes than NMS-3, the ankle shows clear deviations.}
    \label{fig:roms}
\end{figure}

Joint kinematics for all trials are shown in Fig.\,\ref{fig:roms}. The hip kinematics are not changing significantly. When replacing GAS in NMS-1 and NMS-3 only minor changes are visible in knee and ankle comparing NMS-1 to NMS-ref and NMS-3 to NMS-2. Replacing SOL shows impact on ankle and knee kinematics. NMS-2 and NMS-3 exhibit decreased knee flexion in stance and increased knee flexion in swing. The ankle plantar flexion is slightly decreased in stance. In swing the ankle dorsi-flexes rapidly and at $70\%$ stride time hard stops ($20^o$) engage. We assume the spring switch off at the end of stance described in Sec.\,\ref{subsec:mtu_repl} and thus the resulting immediate force drop to zero cause this artifact.\par

\begin{figure}[htb]
    \centering
    \includegraphics[trim=2cm 0.5cm 2cm 0.75cm, clip, width=\textwidth]{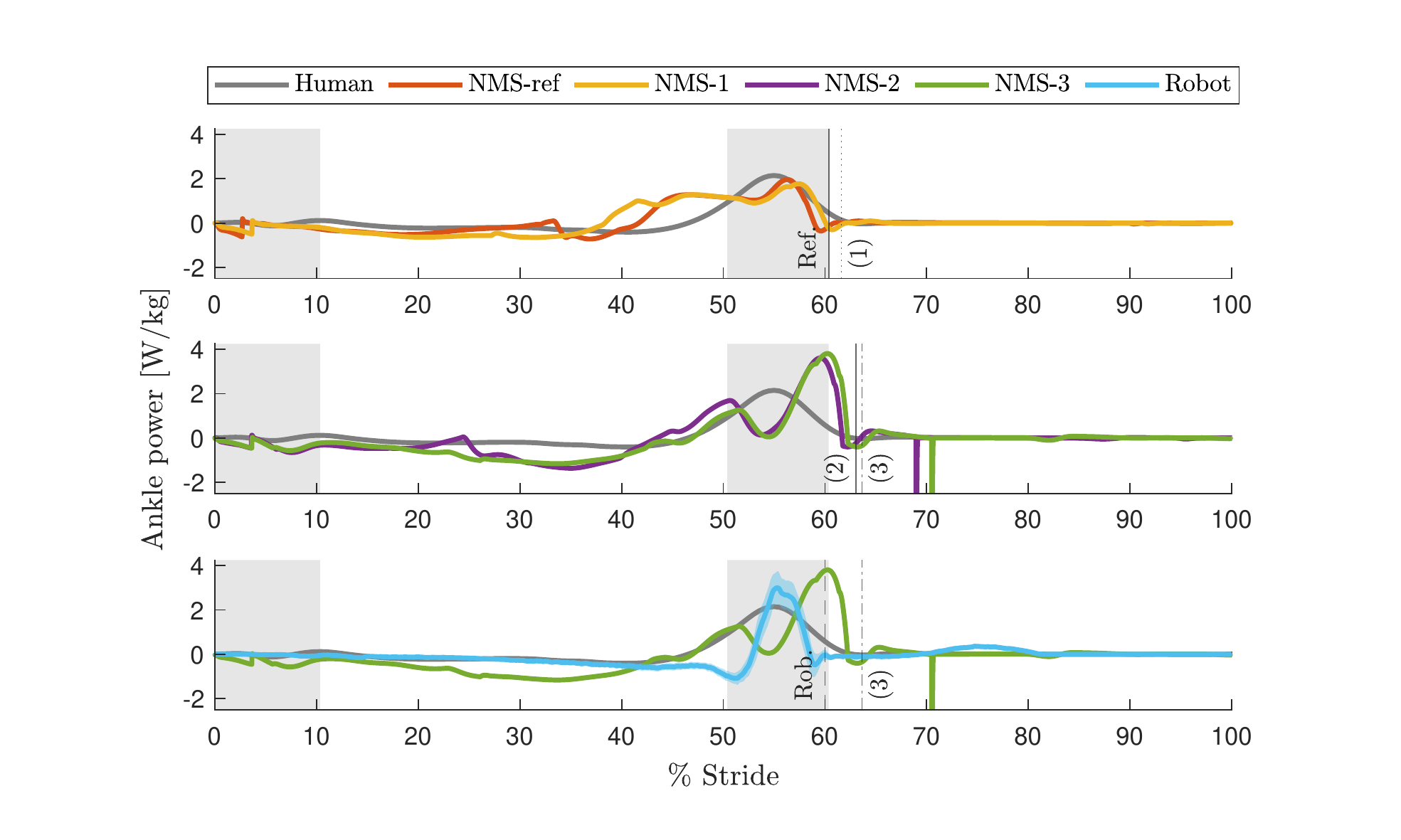}
    \caption[Ankle Joint Power Curve.]{\textbf{Ankle power curves.} Human data from \cite{Lipfert.2014}, sixth step for all NMS trials. Gray areas indicate human double support, vertical lines take off. \textit{Top:} NMS-ref and NMS-1. Ankle joint unloading starts earlier in NMS than in humans, for NMS-1 slightly earlier than for NMS-ref. \textit{Middle:} NMS-2 and NMS-3. Loading with higher negative power compared to human data and faster unloading with higher positive power peak compared to NMS-1 and NMS-ref. Around 70\% stride ankle hard stops engage. \textit{Bottom:} NMS-3 and robotic data. The maximum peak for the robot is higher than for humans with similar peak timing but faster unloading. NMS-1,2,3 show a later take off than NMS-ref and humans.}
    \label{fig:Power_Curve}
\end{figure}

All trials show the characteristic APO (Fig.\,\ref{fig:Power_Curve}). In NMS-ref and NMS-1 the APO starts earlier than in humans ($37\%$ stride for NMS-ref, $40\%$ for NMS-1 vs. $45\%$ for human). In NMS-2 and NMS-3 the APO starts similar to human data and shows a double hump profile with interrupted unloading during the first half of the final double support phase. The total positive power output during APO with \SI{15.9}{J} at \SI{80}{kg} for NMS-ref is comparable to human data with \SI{14.6}{J} at \SI{1.3}{m/s} from \cite{Lipfert.2014} at \SI{70.9}{kg} body mass. The power output increases for all experiments compared to NMS-ref by max. $16\%$ in NMS-2. Negative power for loading and net energy at the ankle decrease from NMS-1 to NMS-3. We see a 1.8 times higher positive power peak when SOL's MTU is replaced by a spring compared to NMS-ref indicating that ankle peak power and thus power amplification (ratio of positive to negative peak power) are mainly determined by SOL. For more details see Tab.\,\ref{tab:ankle}.

\begin{table*}[htb]
    \centering
    \caption[Global Gait Measures, Metabolic Energy Consumption and Ankle Power Output.]{\textbf{Global gait measures and ankle power analysis}. \textit{GM:} average forward velocity of the upper body segment $v_{HAT}$, stride time $t_{s}$, duty factor $DF$ and metabolic energy $\sum E_{metab}$ for the 6th step.
    \textit{Ankle joint:} ankle power analysis for stance phase. $\Delta E^+$ and $\Delta E^-$ denote the positive and negative power integral over one stride, $\Delta E_{tot}$ is their sum. $P_{Max}$ and $P_{Min}$ are the positive and negative peak values of the joint power. $P_{PA}$ is the power amplification, i.e. the ratio of $\left| ^{P_{Max}}/_{P_{Min}} \right|$. \vspace{0.25cm}
    }
           \setlength{\tabcolsep}{0pt}
        \begin{tabular*}{\textwidth}{@{\extracolsep{\fill}\quad} p{0.25cm} l c r r r r}
        \multicolumn{4}{l}{} & \multicolumn{3}{c}{\textbf{MTU Replacement}}\\\cline{5-7}
        \\[-9pt]
        & & \textbf{Unit} & \textbf{NMS-ref} & \textbf{NMS-1}& \textbf{NMS-2} & \textbf{NMS-3}\\[1pt]
        \hline \\[-9pt]
         & $v_{HAT}$        & $[m/s]$   & 1.3  & 1.3      & 1.3      & 1.2 \\ 
         & $t_s$            & [s]       & 1.2  & 1.2      & 1.2      & 1.1 \\ 
         & DF               & [-]       & 0.6  & 0.6      & 0.6      & 0.6 \\ 
          %
          \multirow{-4}{*}{\begin{sideways} GM 
          \end{sideways}}  
          %
       & $\sum E_{metab}$ & [W/kg]      & 5.9 & 5.5    & 5.3    & 4.7 \\ [1pt]
      \hline \\[-9pt]
      & $\Delta E^+$ & [J] & 15.9 & 18.6 & 18.8 & 17.8      \\ 
      & $\Delta E^-$ & [J] & -10.8 & -13.1 & -21.3 & -23.1   \\ 
      & $\Delta E_{tot}$ & [J] & 5.2 & 5.5 & -2.5 & -5.3     \\[1pt] \cline{2-7} 
      \\[-9pt]
      & $P_{Max}$ & [W] & 157.5 & 142.0 & 288.8 & 304.7     \\ 
      & $P_{Min}$ & [W] & -57.3 & -51.8 & -109.4 & -93.0    \\ 
    \multirow{-6}{*}{\begin{sideways} Ankle joint \end{sideways}} 
      & $P_{PA}$ & [-] & 2.8 & 2.7 & 2.6 & 3.2            \\ [1pt]
        \hline 
        \end{tabular*}
    \label{tab:ankle}
\end{table*}

\subsection{Bipedal Robot}
We successfully implemented NMS-3 on the bipedal robot. The robot walks continuously with a purely passive ankle joint at \SI{0.55}{m/s} on a treadmill. With a leg length of \SI{350}{mm} the robot walks at a \textit{Froude number} of 0.09 compared to 0.16 for the NMS. 
Joint kinematics and ankle power curve of the robot, human, and NMS-3 are shown in Fig.\,\ref{fig:roms} and \ref{fig:Power_Curve}. Hip and knee trajectories actively controlled by CPGs match NMS data reasonable well. Ankle kinematics however show clear differences. Especially dorsiflexion in swing and at the beginning of stance is missing.
The ankle power curve shows the characteristic APO. With a positive peak power of $P_{max} = 3.0\,W/kg$ and a negative peak power of $P_{min} = -1.1\,W/kg$ the power amplification ratio $P_{PA} = 2.7$ is similar compared to NMS-ref, but smaller than in NMS-3 (Tab.\,\ref{tab:ankle}). The peak timing at $55\%$ stride is comparable to human data, but the unloading happens faster and at a higher peak power. \textit{For detailed robotic results see \cite{Kiss.2022}}. 

\section{Discussion}
Our initial research questions were \textbf{(I) Do humans need active ankle actuation for steady state  walking} and \textbf{(II) What are the individual contributions of SOL's and GAS's MTUs to the APO in human walking.}\par 
We used a 2D neuromuscular simulation and replaced (1)\,GAS, (2)\,SOL and (3)\,GAS\,and\,SOL MTUs by linear springs. We validated NMS-3 on a bipedal robot with active hips and knees and passive ankle joints.\par
%
All NMS trials and the robot walked stable showing an impulsive APO. With use of only elastic energy for ankle plantar flexion in NMS-3 and robot, continuous walking was possible supporting the hypothesis that humans mostly use the passive-elastic properties of GAS and SOL MTUs for walking \cite{Cronin.2013, Lichtwark.2007}.\\
In simulation we could reduce the metabolic costs by 20\% from $5.9\,W/kg$ in the NMS-ref to $4.7\,W/kg$ in NMS-3. Comparable metabolic costs as in NMS-3 were observed for humans with $5-6\,W/kg$ \cite[Figure 4(b)]{Umberger.2010}. Ankle-powered gait is known to be very efficient \cite{Zelik.2014} as efficiency of muscle work increases from proximal to distal \cite{Sawicki.2009}. Our results show, that therefore the APO and involved MTUs play a crucial role for the energy efficiency of human walking.\par

SOL is the stronger muscle with $F_{max,SOL}=4000N$ compared to $F_{max,GAS}=1500N$ \cite{Yamaguchi.1990}, allowing for more elastic energy to be stored in the Achilles tendon. SOL moreover shows a larger influence on the reduction of metabolic energy than GAS (Tab.\,\ref{tab:ankle}). SOL's strong influence on metabolic measures is in line with findings from \textcite{Collins.2015} where an unpowered exoskeleton with a spring in parallel to the Achilles tendon reduced the metabolic energy consumption mainly by decreasing SOL's activity.\\
GAS shows less influence on the global joint kinematics when its MTU is replaced by a spring. Biarticular muscles have long tendons to support rapid recoil and the use of elastic energy \cite{Schumacher.2020} thus the MTUs behavior has a stronger elastic component.\\ 
Both muscles, GAS and SOL, have a passive role in human APO, isometrically holding the force when elastically loading the Achilles tendon \cite{Ishikawa.2005}. Thus their MTUs can be replaced by fully passive springs as shown here. Further research is required to investigate the role of SOL and GAS in more detail, e.g. by eliminating GAS and/or SOL in the NMS, varying the stiffnesses for spring replacements in NMS-1,2,3 and investigating the power flow between knee and ankle.\par
NMS and robot only resemble sagittal plane motion and we used a coarse fitting for the spring replacement. Stance phase joint kinematics show good agreement with human data. Swing kinematics however do not match, especially for NMS-2 and NMS-3. The robot shows clear deviations from human data regarding ankle kinematics. Especially improvements on dorsiflexion capability are needed.\par
%
The well-known ankle power curve and the impulsive APO are preserved in both, robot and NMS. However, when SOL's MTU is replaced by a spring in NMS-2 and NMS-3, the APO comes with interrupted unloading; a case which has not been tested in the robotic model and for which no human data exists. We suspect the foot model has a yet unclear influence on this phenomenon. To investigate the foot model's influence in more detail we plan further studies with simulation and robotic hardware.\par 

We conclude, that (I) steady state walking with impulsive APO does not require active ankle actuation. Elastic structures to store and release energy in a catapult-like fashion are sufficient in simulation and on a bipedal robot. For matching other gait phases or different gaits however, active muscles or more complex passive structures may be necessary. \\
(II) SOL is responsible for providing the necessary peak force and influences the APOs power amplification. GAS is responsible for coordination and energy transfer between knee and ankle using the elastic properties of its long tendons. Both muscles mainly hold the forces arising during elastic loading of their MTUs without active contraction. \par
Our research shows, that simulation in combination with a robotic model can help to improve the understanding of biomechanics and control principles in human walking. Simulation and robot facilitate decomposing complex structures and verifying findings under real-world conditions. \par
For designing prostheses, exoskeletons, or humanoid robots we conclude that no complex ankle actuation is needed to functionally reproduce the governing dynamic behavior of the ankle during stance. Incorporating more mechanical intelligence at the ankle could be a promising approach to restore and replicate human lower limb function without complex control or actuation to improve gait efficiency and reduced design complexity. 

\section{Acknowledgements}
The authors thank Emre Cemal Gonen and An Mo for their help with the robot data. The project is funded by the Deutsche Forschungsgemeinschaft (DFG, German Research Foundation) - 449427815.

\footnotesize{%
    \printbibliography%
}%

\end{document}